\documentclass[runningheads]{llncs}

\usepackage[T1]{fontenc}

\usepackage{graphicx}
\usepackage{subcaption}
\usepackage{booktabs}
\usepackage[bb=boondox]{mathalfa}
\usepackage{comment}
\usepackage{amsmath,amssymb}
\usepackage{color}
\usepackage[hyphens]{url}
\usepackage[breaklinks]{hyperref}
\usepackage{orcidlink} 
\renewcommand{\orcidID}[1]{\orcidlink{#1}}

\usepackage[table]{xcolor}      
\newcolumntype{g}{>{\columncolor{gray!15}}c}
\usepackage[inline]{enumitem} 

\newif\ifreview
\reviewfalse

\ifreview
	\usepackage{lineno}

	\linenumbers
\fi

\begin{document}


\def\SubNumber{042}

\def\GCPRTrack{Special Track: Pattern recognition in the life and natural sciences}

\title{Assessing Foundation Models for Mold Colony Detection
with Limited Training Data}

\ifreview
\else
	\titlerunning{Assessing Foundation Models for Mold Colony Detection}

	\author{Henrik Pichler\inst{1,2}\orcidID{0009-0008-6616-3108} \and
	Janis Keuper\inst{1}\orcidID{0000-0002-1327-1243} \and
	Matthew Copping\inst{2}\orcidID{0009-0002-0844-4055}}
	
	\authorrunning{H. Pichler et al.}
	
	\institute{Offenburg University of Applied Sciences, Offenburg, Germany \\\email{janis.keuper@hs-offenburg.de} \and BioStates GmbH, Baden-Baden, Germany \\\email{h.pichler@biostates.de}}
\fi

\maketitle              

\begin{abstract}
The process of quantifying mold colonies on Petri dish samples is of critical importance for the assessment of indoor air quality, as high colony counts can indicate potential health risks and deficiencies in ventilation systems. Conventionally the automation of such a labor-intensive process, as well as other tasks in microbiology, relies on the manual annotation of large datasets and the subsequent extensive training of models like YoloV9. To demonstrate that exhaustive annotation is not a prerequisite anymore when tackling a new vision task, we compile a representative dataset of 5000 Petri dish images annotated with bounding boxes, simulating both a traditional data collection approach as well as few-shot and low-shot scenarios with well curated subsets with instance-level masks. We benchmark three vision foundation models against traditional baselines on task specific metrics, reflecting realistic real-world requirements. Notably, MaskDINO attains near-parity with an extensively trained YoloV9 model while finetuned only on 150 images, retaining competitive performance with as few as 25 images, still being reliable on $\approx{70\%}$ of the samples. Our results show, that data-efficient foundation models can match traditional approaches with only a fraction of the required data, enabling earlier development and faster iterative improvement of automated microbiological systems with a superior upper-bound performance than traditional models would achieve.

\end{abstract}
\section{Introduction}

\subsection{Problem definition}
Maintaining clean air in office and production environments is of critical importance to ensure the health and well-being of employees. In accordance with VDI Guideline 6022 \cite{VDI6022}, the quality of air entering premises via ventilation must be maintained, as poor air quality results in health issues such as respiratory problems. Mold spores represent a significant source of indoor air pollution, which underscores the need for regular monitoring and assessment of air quality. This is particularly important in settings with sensitive environmental conditions, such as hospitals and laboratories \cite{Measuring_Mold,care_for_your_air_2024}.

Conventionally air quality assessment involves the collection of air or surface samples on Petri dishes, subsequent incubation for several days to permit the growth of mold colonies, and the enumeration of these colonies. Although the counting process is relatively straightforward, it is inherently labor-intensive due to the manual effort required. In addition, the air quality assessment often requires the differentiation of the colonies as well. This complicates the task, as often a microscopic evaluation is required in this case.
In consideration of the growing demand for efficient air quality monitoring \cite{Air_quality_market_size}, there is a pressing need for automated solutions that can reduce manual workload and save time without compromising accuracy.

\subsection{Training from scratch on new tasks}
When training a machine learning model on a new task, a standard approach would often be to first collect large amounts of the data needed for solving the task. In the past, this was necessary, as models like AlexNet, VGG and ResNet needed a lot of data to perform at an acceptable level \cite{resnet,AlexNet,vgg}. This resulted in the development of large datasets for different tasks, including the ImageNet dataset \cite{ImageNet} for image classification (1.2–14 million images) and the MS-COCO dataset \cite{CoCo} for object detection (over 200 thousand labeled images), for which the annotation was especially labor-intensive. Even with these datasets, models were dataset-centric, needing a lot more data to potentially adapt to new tasks while still being biased and error-prone to changes in the data \cite{posebias,human_ai_comp,texturebias,CLIP_web}. 
So even though there is an abundance of large annotated datasets, adapting to new tasks takes time and annotation effort.

\subsection{Foundation Models in Computer Vision}
Recently foundation models have emerged as a prominent topic of interest in both the natural language processing and computer vision domains. These models employ extensively trained backbones, allowing them to capture general, robust and transferable representations of concepts within images. This enables them to perform well on a variety of tasks with minimal or no additional training, known as zero-shot or few-shot learning \cite{Zero-Shot}. While a foundation model is often described as a model trained in an unsupervised or self-supervised manner \cite{Foundation_Models}, studies show that backbones trained with a supervised objective are on par with self-supervised backbones. Often they even excel their counterpart, e.g. when the supervised dataset is larger than the unsupervised \cite{BoB} or when unsupervised pretraining objectives are combined with supervised ones \cite{supmae}. 

Despite their wide-ranging applicability, the use of foundation models in niche domains, such as the automated enumeration of mold colonies, remains underexplored. The majority of existing research and applications concentrate on general-purpose datasets such as ImageNet \cite{ImageNet}, COCO \cite{CoCo} or similar ones, which feature everyday objects and scenes and lack the specific characteristics present in specialized domains. This gap in knowledge makes it unclear whether foundation models can generalize effectively to such tasks. \cite{objectnet} shows how varying the viewpoint of everyday objects poses a significant challenge for many models that previously performed well on datasets like ImageNet. This is to some degree still the case, even when looking at the leading foundation model for image classification, CoCa. Though it has a 91\% accuracy on the ImageNet dataset, its performance drops by almost 10\% on the ObjectNet dataset \cite{objectnet,coca_paperswithcode}. This effect might be intensified when switching to a completely new domain. 
Nevertheless, given their inherent strengths in vision tasks, foundation models present a promising base for adapting to new domains with relatively small amounts of data. This study aims to bridge this gap by evaluating the performance of foundation models in the novel domain of automated mold colony counting, thereby providing insights into their capabilities and limitations when applied to specialized tasks with limited training data, as is the case for many microbiological applications.

\subsection{Goal}

In this study, we aim to demonstrate that extensive data collection is not a prerequisite for effective model training and utilization. We seek to illustrate that foundation models can rapidly learn and accurately perform the task of mold colony counting, significantly reducing the time and resources required to deploy automated solutions. 
The contributions of this study can be summarized as follows. \textbf{(i)} We conduct the first systematic benchmark of microbiological colony counting with vision foundation models, namely MaskDINO \cite{MaskDINO}, SAM-2 \cite{SAM2}, and RF-DETR \cite{rf-detr}, under carefully annotated high-, few- and low-shot data regimes (see Sec. \ref{model_results}). \textbf{(ii)} We introduce task-oriented counting metrics, aligning with realistic laboratory reporting requirements and complement average precision (see Sec. \ref{eval_metrics}). \textbf{(iii)} We compile a 5,000-image Petri dish dataset with bounding box and instance segmentation mask annotations, including stratified subsets for few-shot studies (see Sec. \ref{dataset_creation}). We therefore aim to provide practical guidelines for early-stage model deployment in specialized computer vision domains, where data is scarce (see Sec. \ref{discussion}).
The study could enhance the efficiency of air quality monitoring without the need for large, annotated datasets typically required by traditional methods. In return, the automation of the task could lead to reduced sample evaluation costs, enabling customers to submit more samples or submit samples more frequently, resulting in cleaner air. Furthermore, the insights derived from this study can be applied to other microbiological tasks, underscoring the potential of foundation models to propel specialized research areas despite limited data availability. To our knowledge, this is the first work assessing the low data performance of foundation models in the domain of microbiological colony detection.

\section{Related work}

\subsection{Deep Learning applications in microbiology}

The field of microbiology has seen a growing reliance on deep learning for tasks such as species identification, colony enumeration, and early-stage detection of pathogens. However, the acquisition of large, annotated datasets remains a significant challenge in many microbiological applications. Lab-based image collection involves the culturing and precise imaging of samples, while the heterogeneity of data, including different species, growth phases, and plate types, further complicates model training.
\cite{Early_detection} developed a CNN-based system that could classify three bacterial types on agar plates, correctly identifying approximately 80\% of colonies 12 hours earlier than standard protocols. However, their approach required high-quality images and was limited to one plate at a time, highlighting scalability issues for high-throughput laboratories. \cite{aspergillus_classification} employed a CNN to classify Aspergillus species, achieving F1-scores above 95\%, but only after training on thousands of carefully prepared images, primarily featuring a single, unmistakable colony. Such data requirements can be prohibitive in real-world scenarios where multiple overlapping colonies often appear.
In contrast, \cite{33_bacteria_microscope} reported a 97\% accuracy in bacterial classification, but their training set contained only 10 images per species. This limitation highlights the risk of overfitting, which can be difficult to detect in limited data sets. \cite{89_mold_microscope} attempted an expanded classification of 89 different fungi species. However, the model's accuracy was modest, with approximately 65\% when presented with average-quality microscopic images, underscoring the potential for performance decline in less controlled settings.

Recent studies have emerged to advocate for the adoption of advanced architectures such as ConvNeXt \cite{ConvNeXt}, which effectively handle increased variability and smaller datasets with greater efficacy than their predecessors. For instance, \cite{ConvNeXt_macular} report accuracy gains of up to 10–15\% when comparing ConvNeXt to conventional CNNs in the context of the classification of macular degeneration, suggesting that contemporary models may offer enhanced scalability and robustness, even in scenarios where data is limited.

\cite{med_class_tl} suggests that research in microbiological machine learning tasks remains dependent on legacy CNN backbones such as InceptionNet-v3 \cite{inceptionnet} and ResNet-50 \cite{resnet}, yet achieves competitive outcomes through transfer learning rather than through training from scratch. A systematic review of 121 medical-imaging papers reveals that InceptionNet-V3 was the most prevalent model, while fine-tuned feature-extractor regimes exhibited superiority over fully re-trained networks, thereby substantiating the viability of older architectures when strong priors are imported from ImageNet or analogous sources. In consideration of the microbiology-specific studies previously outlined, these observations indicate that architectures explicitly designed for adaptation, i.e., modern foundation models, should exhibit an even better capacity to manage small, heterogeneous datasets and rapidly generalize to novel tasks.

These studies underscore the pressing need to apply deep learning to a broader range of microbiological tasks and reveal how limited, specialized datasets often constrain generalization, with scalability issues becoming more evident when image conditions degrade or when more diverse samples are introduced. Motivated by these insights, this study investigates whether modern model architectures, especially foundation models, can achieve a strong performance in mold colony counting without requiring large-scale datasets.

\subsection{Object Detection and Segmentation Architectures}

Classical object detection and segmentation models, such as Yolo \cite{yolo-overview,Yolov9} for object detection and Mask R-CNN \cite{MaskRCNN} for segmentation, have long served as robust baselines in computer vision tasks. These approaches typically rely on training with large amounts of annotated data and benefit from pretrained backbones on large-scale datasets like ImageNet \cite{ImageNet} or MS COCO \cite{CoCo}. This extensive supervised pretraining often enables them to achieve strong performance in well-studied domains.

\section{Methodology}

\subsection{Dataset Creation}
\label{dataset_creation}
To evaluate the capability of foundation models in the task of mold colony counting, a dataset that represented the complexity of the problem while accounting for different data availability scenarios was essential. Consequently, a dataset comprising three scenarios was developed: one intended for traditional object detection models trained with a substantial amount of data, and two others designed for testing the low-data capabilities of foundation models. To ensure a valid and reliable comparison of the models, a validation and a test dataset were constructed.

\subsubsection{Data Collection}
High resolution images (1400x1400 pixels) of Petri dishes with mold colonies were gathered after an incubation period. The surface samples were obtained from a variety of locations throughout Germany and cultivated in DG18-Agar based Petri dishes, enabling only the growth of mold colonies. Most samples showed the growth of at least one colony. A standardized setup with a high-resolution camera and consistent lighting from the side was used to capture images from the obtained and incubated samples. This setup reflected realistic conditions relevant to mold colony counting.

\begin{figure}
    \centering
    \begin{subfigure}{0.32\linewidth}
        \includegraphics[width=\linewidth]{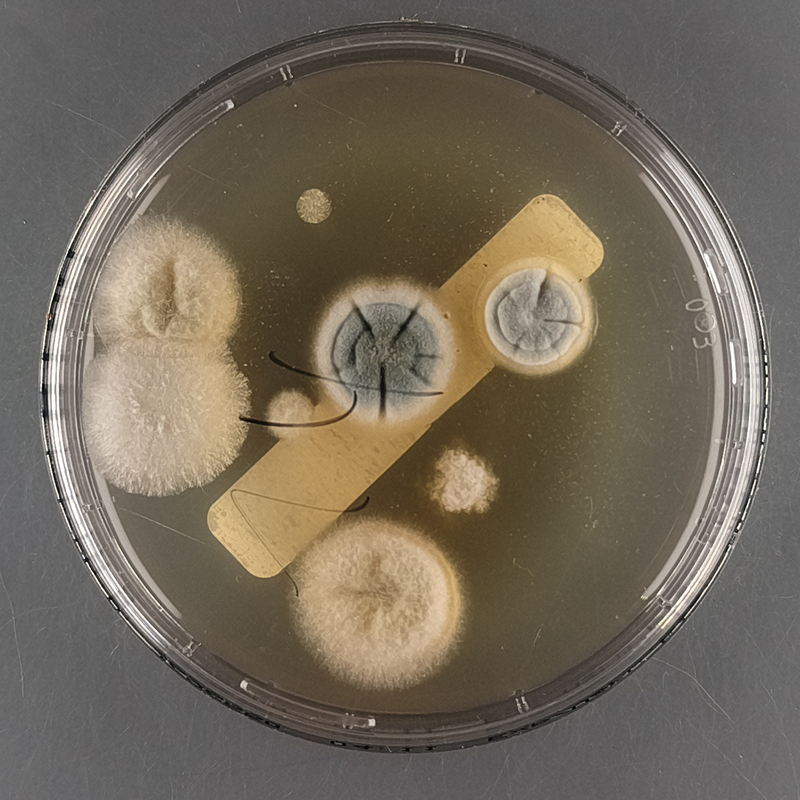}
        \caption{}
        \label{fig:normal_plate}
    \end{subfigure}
    \hfill
    \begin{subfigure}{0.32\linewidth}
        \includegraphics[width=\linewidth]{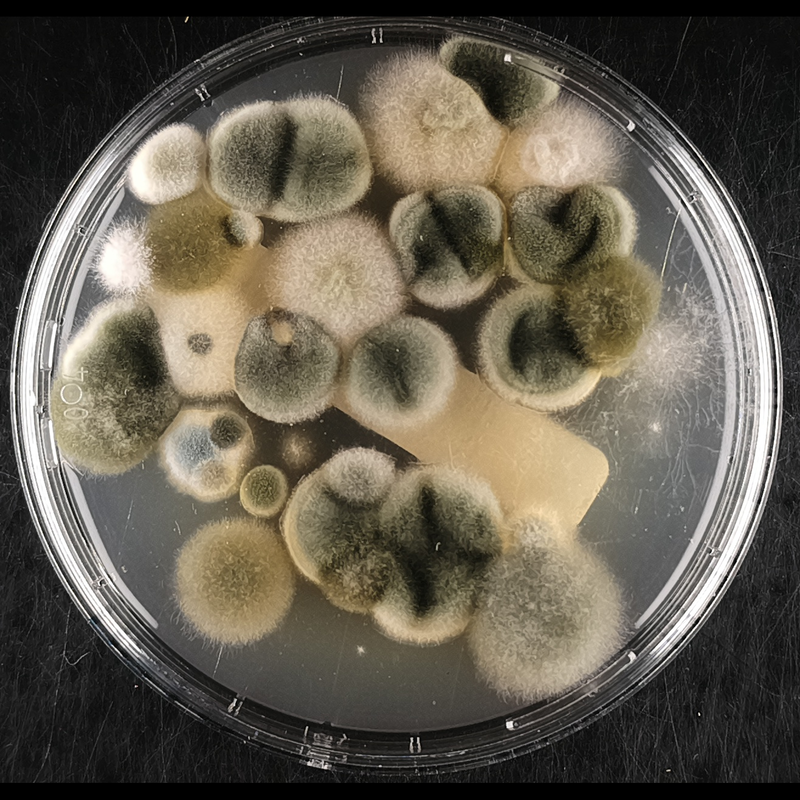}
        \caption{}
        \label{fig:normal_plate_overlapping}
    \end{subfigure}
    \hfill
    \begin{subfigure}{0.32\linewidth}
        \includegraphics[width=\linewidth]{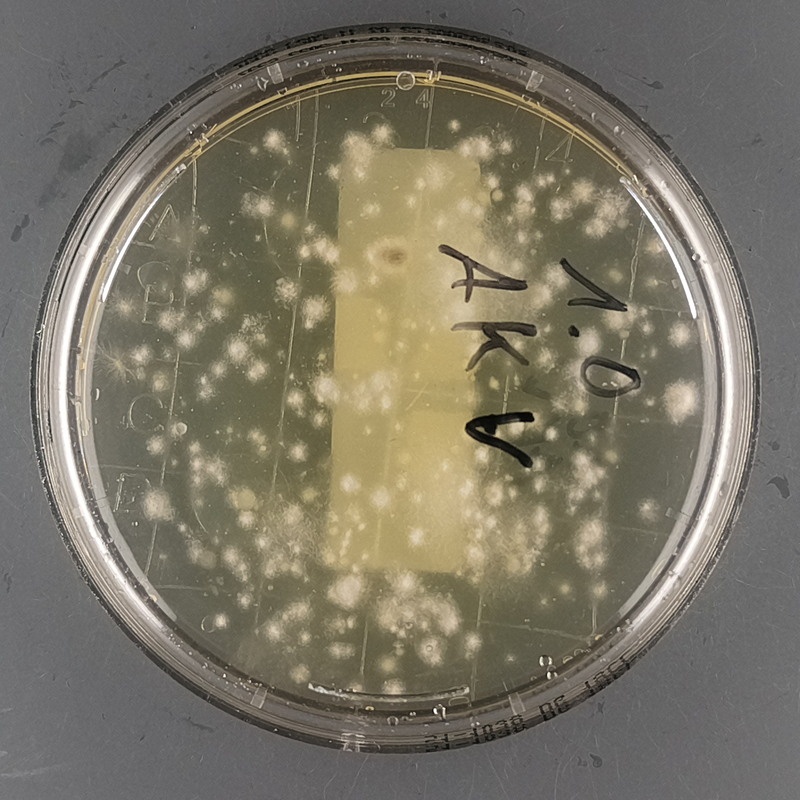}
        \caption{}
        \label{fig:plate_small_colonies}
    \end{subfigure}
    \caption{Comparison between different mold colony growth patterns. (a) A typical sample with well-distinguishable colonies, (b) well-distinguishable but overlapping colonies on a darker background (caused by a changed backdrop throughout the capturing process), and (c) many small colonies, presenting a more challenging scenario.}
\end{figure}

\subsubsection{Annotation}
The data collection resulted in 5,000 manually annotated images, each containing at least one mold colony. The dataset was divided into three subsets for the training, validation, and testing. The ratio of images in the subsets was 4,000:500:500. To simulate low-data scenarios, a subset of 150 training images was selected from the large 4,000-images dataset. To simulate scenarios with even less data, e.g. when data collection is generally difficult or was just started, 6 additional 25-image subsets were sampled from the 150-image subset, using the Stratified K-Fold method based on the number of instances per image \cite{StratifiedKFold}. The usage of multiple 25-image subsets ensured that the results were not a random success or failure, but could be reliably evaluated. The resulting images as well as the validation and test set were further annotated with instance segmentation masks, ensuring a proper comparison between instance segmentation and object detection methods.

\subsection{Experimental setup}
\label{setup}
The goal was to provide insights into modern segmentation/foundation model capabilities by comparing them to more traditional methods. Since accurate colony counts were important and colonies often overlap (see Fig. ~\ref{fig:normal_plate_overlapping}), an object detection and instance segmentation approach was chosen for this study, allowing for a more precise determination of colony boundaries than e.g. a semantic segmentation approach would be able to do. The models were selected for their predefined training scripts, which simplified adaptation to the task. The objective was to understand each approach's strengths, limitations, and capabilities under various data conditions, rather than developing a novel model. Practically, minimal model adaptation with limited fine-tuning is desirable for effective real-world applications.

The comparison covered two conventional architectures (YoloV9 \cite{Yolov9} and Mask-RCNN \cite{MaskRCNN}) and four foundation model variants with diverse backbones. Because annotating pixel-accurate masks for the entire 4,000-image dataset is prohibitively time-consuming, only YoloV9 and RF-DETR \cite{rf-detr} were trained on the full dataset with bounding-box annotations to additionally enable a comparison of foundation models for the task of object detection. Mask-RCNN, MaskDINO and its variants \cite{MaskDINO}, and SAM2 \cite{SAM2} were instead trained on smaller subsets that included instance segmentation masks, capturing pixel-level boundaries and separating overlapping colonies more reliably than boxes alone. All models were evaluated on the same 500-image test set, either using bounding boxes or segmentation masks, based on their training regime. A detailed summarization of the models can be found in Tab. ~\ref{tab:model_overview}. An in-depth overview of the training parameters can be found in the supplementary material, but it can be generally said that all models were trained mostly with their standard training parameters until no further improvements could be seen.

SAM2 is a prompt-based segmentation model which can be used to automatically generate masks with a grid of points as input and postprocessing the output with quality filtering methods and de-duplication using non-maxima suppression. The models' results depend on the parameters used for the automatic mask generation, so an optimization of the parameters was conducted, including a custom grid to guide the model towards better predictions. See supplementary material for additional information.

\begin{table}[!htbp]
\centering
\caption{Overview of all models evaluated in this study.}
\label{tab:model_overview}
\begin{tabular}{l | c | c | c | c }
\toprule
\textbf{Model} & \textbf{Backbone} & \textbf{Pre-training} & \textbf{Fine-tuning data} & \textbf{\# Params}\\
\midrule
YoloV9-E & - & MS COCO & 4k / 150 / 25 bbs & 57.3 M\\
RF-DETR-L  & DinoV2-b & ImageNet-21k & 4k / 150 / 25 bbs & 135 M\\
Mask R-CNN  & R50 & MS COCO & 150 / 25 masks & 45.9 M\\
MaskDINO & R50 & MS COCO & 150 / 25 masks & 52 M\\
MaskDINO  & Swin-L & ImageNet-21k & 150 / 25 masks & 223 M\\
SAM2-base  & Hiera & SA-1B\,+SA-V & 150 / 25 masks & 80.8 M\\
\bottomrule
\end{tabular}
\end{table}

\subsection{Evaluation Metrics}
\label{eval_metrics}

In order to assess the performance of the models under varying data regimes, a set of metrics was employed for the purpose of evaluating both the precision of the segmentation and the accuracy of the quantification of mold colonies.

\subsubsection{Segmentation and Detection Metrics}
The evaluation considered the outputs of each model. YoloV9 and RF-DETR were evaluated based on the bounding boxes they produced, and all other models were evaluated on the segmentation masks they provided. These outputs were assessed using standard metrics:
\begin{itemize}
    \item \texttt{Mask Average Precision ($AP_{mask}$):} Evaluates overlap between predicted and ground-truth masks across multiple IoU thresholds.

    \item \texttt{Box Average Precision ($AP_{box}$):} Measures the precision of bounding box predictions at varying IoU thresholds.

\end{itemize}

\subsubsection{Quantification Accuracy Metrics}
Since the primary goal is accurate mold colony quantification, a metric was used to assess the models' ability to predict the number of colonies.This metric, Counting Accuracy (CA), describes the percentage of test images with the exact quantity of mold colonies correctly identified:

\begin{equation}
  \text{CA} = \frac{\text{Number of Images with Correct Count}}{\text{Total Number of Images}} \times 100  
\end{equation}

Manual enumeration of mold colonies is subject to small discrepancies. An exact-match metric like counting accuracy can be overly strict for practical purposes. The "Counting Accuracy at 10\% tolerance" (CA@10) extension to counting accuracy considers a prediction "correct" if its absolute relative error with respect to the ground-truth count is at most 10\%, and therefore 90\% correct. Formally,

\begin{equation}
  \text{CA@10} \;=\;
  \frac{1}{N}
  \displaystyle
    \sum_{i=1}^{N}
      \mathbb{1}
      \!\left(
        \frac{\lvert c_i -\hat{c}_i \rvert}{c_i}
        \le 0.10
      \right)%
  \times 100 
  \label{eq:ca10}
\end{equation}

where \(N\) is the number of samples, \(c_i\) is the ground-truth colony count for sample \(i\), \(\hat{c}_i\) is the model’s predicted count, and \(\mathbb{1}(\cdot)\) is the indicator function that returns \(1\) when the condition is satisfied and \(0\) otherwise.  

CA@10 therefore measures the percentage of plates for which the model’s estimate deviates by no more than 10\% from the reference count.  
In routine microbiological quality control, such an error band is acceptable because most of the time it doesn't change the qualitative assessment of contamination level. Therefore, this metric is the most impactful for the study.

To assess the relative counting error of the models, the Mean Absolute Percentage Error (MAPE) in colony count per image (eq. ~\ref{eq2}) was additionally computed.

\begin{equation}
\label{eq2}
    \text{MAPE} = \frac{1}{N} \sum_{i=1}^N \frac{\lvert c_i - \hat{c}_i \rvert}{c_i} \times 100
\end{equation}

\section{Results}

\subsection{Dataset statistics}
A total of 5,000 images of Petri dishes containing 109,636 mold colonies were captured. The dataset was split into three parts: a 4,000-image training set and two 500-image validation and test sets. From the 4,000-image training set, a 150-image subset was sampled, from which six subsets of 25 images each were sampled. A detailed analysis of the distribution of mold colony instances across these dataset partitions reveals a consistent pattern, with median values ranging from five to eight colonies per image and a 90th percentile of around 58 colonies per image. In extreme cases, the number of colonies on a single sample ranged from 327 colonies in the 150-image subset to 424 in the full 4,000-image training set. The smaller 25-image subsets exhibited greater variability, with an average maximum of approximately 166 colonies per image and substantial deviation ($\pm89.53$), thereby underscoring the inherent randomness of sampling. With respect to the relative size of mold colonies, the median colony sizes were consistently small, approximately 0.19\% of the total image area, across most subsets. The majority of colonies did not exceed approximately 1.5\% of the image area, as indicated by the 90th percentile values (ranging from 0.88\% to 1.57\%). It is noteworthy that remarkably larger colonies were uncommon. Nevertheless, when present in an image, they could account for up to 70.1\% of the image area in the complete dataset.

Additional details about the dataset can be found in the supplementary material.

\subsection{Model training}

\noindent44 model trainings were carried out. Additionally to their regular trainings, each model was trained on all 6 25-image subsets (see Tab. ~\ref{tab:model_overview}), for which the results were aggregated. All models were tested on the same test set with 500 images using the metrics denoted in Sec. \ref{eval_metrics}. 

\label{model_results}
\begin{table}[h]
    \centering
    \caption{Model results, where the results for the models trained on the 25-image splits are shown as the mean $\pm$ standard deviation. The CA@10 metric (highlighted in gray) is the most insightful metric here, showing how well the model performs in quantification, while still being reliable.}
    \begin{tabular}{l | l | g | c | c | c}
        \toprule
        \multicolumn{1}{c|}{\textbf{\shortstack{Train\\Size}}} 
         & \multicolumn{1}{c|}{\textbf{Model}} 
         & \textbf{CA@10} $\uparrow$ 
         & \textbf{AP} $\uparrow$ 
         & \textbf{MAPE} $\downarrow$ 
         & \textbf{CA} $\uparrow$ \\
        \midrule
        4\,000 & YoloV9-E      & 73   & \textbf{59.89} & 8    & 57.6 \\
               & RF-DETR-L     & \textbf{84.4} & 51.74 & \textbf{5.95} & \textbf{65.8} \\
        \hline
        150   & SAM2-base     & 15   & 22.1  & 62.4 & 14.2 \\
              & Mask R-CNN    & 37   & 46.9  & 45.7 & 23.8 \\
              & YoloV9-E      & 38.4 & 39.44 & 28.2 & 31.8 \\
              & RF-DETR-L     & 69   & 40.65 & 10.89 & 54.2 \\
              & MaskDINO-R50  & 69   & 50.06 & 13   & 51.4 \\
              & MaskDINO-Swin & \textbf{72.6} & \textbf{51.59} & \textbf{8} & \textbf{56.2} \\
        \hline
        25    & SAM2-base     & $10.23 \pm 1.53$ & $16.13 \pm 1.46$ & $108.68 \pm 9.87$ & $7.47 \pm 2.11$ \\
              & YoloV9-E      & $25.67 \pm 6.81$ & $31.95 \pm 1.26$ & $41.06 \pm 4.31$ & $21.37 \pm 3.87$ \\
              & Mask R-CNN    & $35.97 \pm 8.84$ & $39.18 \pm 1.77$ & $41.43 \pm 17.69$ & $24.63 \pm 2.22$ \\
              & RF-DETR-L     & $51.63 \pm 5.74$ & $33.10 \pm 2.39$ & $19.53 \pm 5.71$ & $41.97 \pm 3.12$ \\
              & MaskDINO-R50  & $58.90 \pm 2.29$ & $44.17 \pm 0.71$ & $20.32 \pm 5.58$ & $44.77 \pm 1.17$ \\
              & MaskDINO-Swin & \textbf{67.30 $\pm$ 1.88} & \textbf{46.89 $\pm$ 0.67} & \textbf{10.85 $\pm$ 0.52} & \textbf{50.13 $\pm$ 0.73} \\
        \bottomrule
    \end{tabular}
    \label{tab:model_performance}
\end{table}
\vspace{9pt}

YoloV9 achieves the highest overall AP performance with 59.8\%, although the RF-DETR model outperforms the model in actual quantification of mold colonies with the highest overall CA @ 10 with 84.4\%, establishing a solid foundation for the rest of the results.

The findings show that the AP cannot be taken as a definitive indicator here and that actual performance is strongly dependent on the task at hand. This shows in the results of the MaskRCNN model, having a comparable $AP_{mask}$ to the MaskDINO models. However, its quantification performance falls far behind these models and YoloV9. The counterpart to this is the RF-DETR model, which remains behind YoloV9 by 8.15\% in AP performance, when trained on 4,000 images, but outperforms it by 11.4\% in CA@10 performance. Hence, dense colony enumeration is best evaluated using several metrics, including the presented CA@10, showing what the maximum quantification performance while still being reliable can be. 

At 150 training images, MaskDINO-Swin achieves the highest overall performance with a CA@10 of 72.6\% (see Tab. ~\ref{tab:model_performance}). This performance is notable compared to the YoloV9 model trained on 4,000 images, which only exceeds the achieved CA@10 performance by approximately 0.4\% while MaskDINO requires $\approx4\%$ of the training data. The baseline foundation model, RF-DETR, outperforms this MaskDINO-Swin model by 11.8\% regarding the CA@10, while showing a similar AP performance. When trained on 150 images, RF-DETR is outperformed by MaskDINO-Swin in all metrics while showing on-par performance with MaskDINO-R50. 

\begin{figure}[h]
\includegraphics[width=0.95\textwidth]{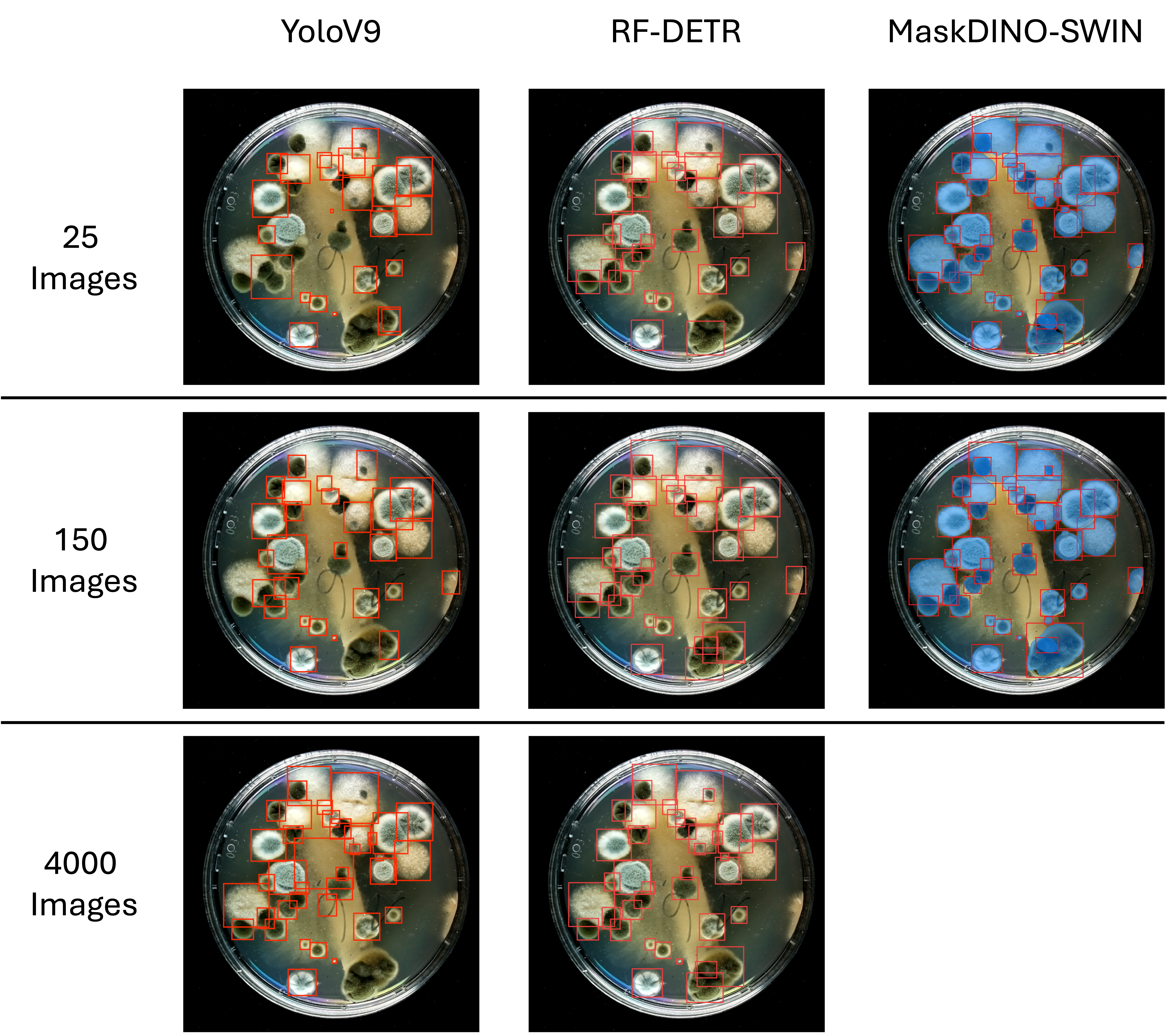}
    \caption{Comparison of predictions made by YoloV9 (left), RF-DETR (middle) and MaskDINO-SWIN (right) trained on different amounts of data. The images show a typical sample with overlapping and small colonies. Additional comparison images can be found in the supplementary material. Images best viewed in color.} \label{fig:pred_comp}
\end{figure}

At 25 training images, MaskDINO-Swin loses $\approx 5\%$ in CA@10, from 72.6\% to 67.3 \%. Still, it is only $\approx5\%$ behind the YoloV9 performance when trained on 4,000 images, while needing just 0.6\% of the training data. Noteworthy, the models performance is roughly three times as high as the YoloV9 performance here. A qualitative investigation shows that YoloV9, even when trained on 4,000 images, misses obvious mold colonies and predicts false positives in background regions. With less data, YoloV9 misses more colonies and makes vague suggestions. MaskDINO clearly separates mold colonies, even overlapping ones, with as few as 25 training images, improving in finer mask predictions with additional training data. RF-DETR also shows strong performance here, with only missing a few colonies when trained on 25 images, improving these mistakes when additional data is added (see Fig. ~\ref{fig:pred_comp}).

SAM2 exhibits a significant performance deficit compared to other models. The model exhibits a performance of 22\% $AP_{mask}$, 14\% CA, and 15\% CA@10 with 150 training images, which are less than one-third of the performance metrics achieved by MaskDINO-Swin in the same regime. A reduction to 25 images results in a significant decline of SAM2 to approximately 16\% $AP_{mask}$ and 10\% CA@10.

\section{Discussion}
\label{discussion}

Though the MaskDINO-SWIN model achieved inferior results when trained on just 25 images, it still achieves a CA@10 result of 67\%, which already is within 6\% of what YoloV9 trained on the full dataset achieved, showing how reliable results can be achieved with just a few carefully annotated samples. This enables a utilization in early stages of projects, while still being reliable for most samples, with an extensive investigation only needed for the most complex samples. The modest accuracy gap between the 25- and 150-image datasets likely stems from the number of samples rather than a shift in data distribution, given the similar statistics of the two datasets. Hence, more samples help the model capture intra-plate variability, but not because they introduce previously unseen colony phenotypes. Therefore, deploying the model in an early stage can help with rapidly increasing the amount of training data, enabling fast improvements. 

Although creating and refining bounding box annotations is faster, using mask annotations might be the ideal method to deploy, especially in low-shot scenarios. This is evident in the comparative performance of the MaskDINO and RF-DETR models. RF-DETR exhibits a 15\% performance deficit when trained on 25 images, indicating it struggles more in extreme low-data conditions than MaskDINO. However, with the full 4,000-image training set, RF-DETR actually surpasses YoloV9 by a large margin, showcasing the benefit of its foundation model backbone at scale. This suggests that a foundation model like MaskDINO might surpass YoloV9 by an even larger gap if more data were available, further enhancing the results.

A steady, monotonic gain is observed across all architectures when the training set increases from 25 to 150 images, signifying that a project may be initiated with a minimal initial sample set and several model candidates to identify a clear leader at start. Subsequent efforts can then be concentrated on this model alone. In the present study, the initial performance of SAM2 indicated that additional training would not be financially viable. MaskDINO and RF-DETR exhibited encouraging upward trends. This iterative workflow method ensures that the initial investment is minimized, ensuring that labeling resources are spent on models that scale effectively with additional data.

In practice, this suggests deploying a comparable application in a new, likely biomedical, computer vision task can be done with a minimal initial effort, when leveraging foundation models, with only a small initial sample set. This initial sample set must be annotated according to the task at hand (e.g., instance segmentation masks instead of bounding boxes for separating overlapping objects). The use of task-specific metrics aligning with real-world scenarios is essential. With regular usage more data can be easily collected, enabling effective scaling with more accurate results. The ability of foundation models to perform well on less complex samples with little training data, allows for  concentrated annotation efforts on more challenging samples, for which the model is likely to struggle with, resulting in a gradual decrease in manual effort and a reduction in the need of supervision.

Overall, foundation models provide two simultaneous benefits when starting a new computer vision project: rapid cold-start performance in few-shot settings and superior upper-bound performance once data accumulates. This makes them the most cost-effective choice across the project life-cycle.

\paragraph{\textbf{Limitations}} SAM2 struggles to segment out negative regions. The prompt generator only provides positive prompts during training, so the model rarely encounters negative regions, never learning to suppress them. This leads to the model misidentifying brown, circular artifacts as colonies. Here, negative samples during training might help, as opposed to merely adding more data.

MaskDINO outperforms RF-DETR in low-shot and few-shot training, though this cannot be reasoned clearly and might be due to the number of parameters, though MaskDINO-R50 has less than half of RF-DETR's, or the annotations. It is clear, that masks capture finer details than bounding boxes, especially for irregularly shaped mold colonies, but further investigation is needed to determine if this gives them an advantage. For this, a model leveraging the same backbone could be trained on bounding boxes and instance segmentation masks, showing the value of investing in pixel-level mask annotation.

Although a standardized imaging setup was employed, the employed side illumination often introduced glare and shadows across mold colonies, complicating the visual separation of adjacent colonies. All reference annotations were produced by a single expert working solely from the captured images without the actual samples at hand, which can increase the risk of wrong labeling. These effects are most pronounced for heavily overgrown samples, where dense colony clusters are hard to properly separate. Future work will therefore incorporate direct plate‑level verification and improved lighting to mitigate these sources of error.

\section{Conclusion}

This study shows that foundation models match or exceed classical detectors in low-data scenarios, demonstrating that data quantity matters not as much than it does with classical approaches when it comes to mold colony counting. These findings suggest an effective workflow: first, focus on annotating a few samples, then refine iteratively to deploy rapidly and grow subsequent datasets.
In a subsequent project, insights from this study will expand to address colony differentiation, a complex but valuable task for automated air quality monitoring. Foundation models may be crucial for rapid system development, especially with the BiomedParse \cite{biomedparse} foundation model, which uses text prompt-based mask generation.

\paragraph{\textbf{Data availability statement}} The dataset that supports the findings of this study is available from the corresponding authors upon request.

%
%
%
%
\bibliographystyle{splncs04}
\bibliography{042-main.bib}

\end{document}